\pdfoutput=1 
\documentclass[preprint]{imsart}
\usepackage{graphicx}
\usepackage{amssymb,amsmath,bm}
\usepackage{amsthm}
\usepackage{subfigure}

\usepackage{natbib}
\def\cite{\citet}

\usepackage[dvips]{geometry}
\startlocaldefs
\newcommand{\trace}{\mathop{\rm tr}\nolimits}
\newcommand{\diag}{\mathop{\rm diag}\nolimits}
\newcommand{\Diag}{\mathop{\rm Diag}\nolimits}
\endlocaldefs

\begin{document}

\begin{frontmatter}
\title{A simple coding for cross-domain matching with dimension reduction via
spectral graph embedding}
\runtitle{cross-domain matching}
\author{\fnms{Hidetoshi}
 \snm{Shimodaira}\corref{}\ead[label=e1]{shimo@sigmath.es.osaka-u.ac.jp}\thanksref{t1}}
\thankstext{t1}{
Supported in part by Grant
KAKENHI (24300106, 26120523) from MEXT of Japan.}
\address{
Division of Mathematical Science\\Graduate School of
 Engineering Science\\
Osaka University\\
 1-3 Machikaneyama-cho\\Toyonaka, Osaka, Japan\\
 \printead{e1}}

\runauthor{H.~SHIMODAIRA}

\begin{abstract}

Data vectors are obtained from multiple domains. They are feature
vectors of images or vector representations of words. Domains may have
different numbers of data vectors with different dimensions.  These data
vectors from multiple domains are projected to a common space by linear
transformations in order to search closely related vectors across
domains. We would like to find projection matrices to minimize distances
between closely related data vectors. This formulation of cross-domain
matching is regarded as an extension of the spectral graph embedding to
multi-domain setting, and it includes several multivariate analysis
methods of statistics such as multiset canonical correlation analysis,
correspondence analysis, and principal component analysis.  Similar
approaches are very popular recently in pattern recognition and vision.
In this paper, instead of proposing a novel method, we will introduce an
embarrassingly simple idea of coding the data vectors for explaining all
the above mentioned approaches. A data vector is concatenated with zero
vectors from all other domains to make an augmented vector. The
cross-domain matching is solved by applying the single-domain version of
spectral graph embedding to these augmented vectors of all the domains.
An interesting connection to the classical associative memory model of
neural networks is also discussed by noticing a coding for
association. A cross-validation method for choosing the dimension of the
common space and a regularization parameter will be discussed in an
illustrative numerical example.

\end{abstract}


\end{frontmatter} 

\section{Introduction} \label{sec:intro}

We consider multiple domains for getting data vectors. Let $D$ be the
number of domains, and $d=1,\ldots,D$ denote each domain. For example,
$d=1$ may be for images, and $d=2$ for words. From domain $d$, we get
data vectors $\bm{x}^d_i \in \mathbb{R}^{p_d}$, $i=1,\ldots,n_d$, where
$n_d$ is the number of data vectors, and $p_d$ is the dimension of the
data vector. They may be image feature vectors for $d=1$, and word
vectors computed by word2vec \citep{mikolov2013distributed} from texts
for $d=2$. Typically, $p_d$ is hundreds, and $n_d$ is thousands to
millions.  We would like to retrieve relevant words from an image query,
and alternatively retrieve images from a word query.

We specify the strength of association between two data vectors
$\bm{x}^d_i$ and $\bm{x}^e_j$ by a matching weight
$w^{de}_{ij}\in\mathbb{R}$ for $d,e=1,\ldots,D$, $i=1,\ldots,n_d$,
$j=1,\ldots,n_e$. (Note that ``matching'' here is nothing related to
that of graph theory.)  We assume the weight is symmetric
$w^{de}_{ij}=w^{ed}_{ji}$. For example, $w^{12}_{11}=3$ for the
association between an image ``apple'' ($\bm{x}^1_1$) and word ``apple''
($\bm{x}^2_1$), and $w^{12}_{12}=1$ for the association between the
image ``apple'' and word ``red'' ($\bm{x}^2_2$). However, it could be
the case that the image apple is unlabeled and $w^{12}_{11}=0$, while
the color may be automatically classified as red and $w^{12}_{12}=1$
remains.  Let $\bar w^{de}_{ij}$ be the matching weight representing the
underlying true associations, and $w^{de}_{ij}$ be observed ones sampled
from the true associations. We assume $w^{de}_{ij}=\bar w^{de}_{ij}$
with a small probability, and $w^{de}_{ij}=0$ otherwise, so that
$\bm{W}^{de}=(w^{de}_{ij})\in\mathbb{R}^{n_d \times n_e}$ would be a
sparse matrix.

The data vectors from all the domains will be projected to a single
common space of $\mathbb{R}^K$ for some $K>0$. Using a matrix
       $\bm{A}^d \in \mathbb{R}^{p_d \times K}$, we define a linear
       transformation by
\begin{equation} \label{eq:yax}
 \bm{y}^d_i = (\bm{A}^d)^T \bm{x}^d_i,\quad
  i=1,\ldots,n_d;\,
  d=1,\ldots,D.
\end{equation}
Here $T$ denotes matrix transpose. Later we use matrix notation such as
$\trace()$ for the matrix trace, $\diag()$ for a diagonal matrix, and
$\Diag()$ for a block diagonal matrix.  Each element of $\bm{y}^d_i\in
\mathbb{R}^K$ is
\[
 (\bm{y}^d_i)_k = (\bm{a}_k^d)^T \bm{x}^d_i,\quad k=1,\ldots,K,
\]       
where $\bm{a}_k^d\in \mathbb{R}^{p_d}$ are defined as
$\bm{A}^d = (\bm{a}_1^d,\ldots,\bm{a}_K^d)$. The error function of
cross-domain matching is
\begin{equation} \label{eq:phia}
 \phi(\bm{A}^1,\ldots,\bm{A}^D) = \frac{1}{2}\sum_{d=1}^D \sum_{e=1}^D \sum_{i=1}^{n_d} \sum_{j=1}^{n_e} w^{de}_{ij} \|\bm{y}^d_i - \bm{y}^e_j\|^2,
\end{equation}
and we would like to find $\bm{A}^1,\ldots,\bm{A}^D$ that minimize
(\ref{eq:phia}) subject to certain constraints. This is a supervised
learning with the matching weights as training data. It handles the
problem of semi-supervised learning and missing observation by simply
letting unobserved weights zero.  For a new query image, say, the data
vector $\bm{x}^1\in\mathbb{R}^{p_1}$ is transformed to $\bm{y}^1 =
(\bm{A}^1)^T \bm{x}^1$. Then look for points close to $\bm{y}^1$ in the
collection $\{\bm{y}^d_i\}$.  By working on the common space in this
way, we will perform data retrieval across domains and data fusion from
multiple domains.

This formulation of cross-domain matching is regarded as an extension of
the spectral graph embedding of \cite{yan2007graph} to the multi-domain
setting, and similar approaches are very popular recently in pattern
recognition and vision \citep{correa2010multi, yuan2011novel,
kan2012multi,huang2013cross,shi2013transfer,
wang2013learning,gong2014multi,yuan2014graph}. In particular, the
formulation reduces to a classical multivariate analysis of statistics,
known as the multiset canonical correlation analysis (MCCA)
\citep{kettenring1971canonical, takane2008regularized,
tenenhaus2011regularized} by letting $n_1=n_2 = \cdots =n_D$ and
connecting all vectors across domains with the same index as
$w^{de}_{ii}\neq 0$ and $w^{de}_{ij}=0$ for $i\neq j$.  Class labels are
coded by indicator variables (called dummy variables in statistics) and
treated as domains; they appear in canonical discriminant analysis and
correspondence analysis. The formulation becomes the classical canonical
correlation analysis (CCA) of \cite{hotelling1936relations} by further
letting $D=2$, or it becomes principal component analysis (PCA) by
letting $p_1=p_2=\cdots=p_D=1$.

In this paper, we do not intend to propose a novel method. Instead, we
will introduce an embarrassingly simple idea of coding the data vectors
for explaining all the above mentioned approaches. This coding is similar
to that of \cite{daume2007frustratingly}. Let $P=\sum_{d=1}^D p_d$ and
$N=\sum_{d=1}^D n_d$. The data vector $\bm{x}^d_i$ is coded as an
augmented vector $\bm{\tilde x}^d_i\in \mathbb{R}^P$ defined as
 \begin{equation} \label{eq:coding}
  ( \bm{\tilde x}^d_i )^T = \Bigl( (\bm{0}_{p_1})^T,\ldots,
   (\bm{0}_{p_{d-1}})^T, (\bm{x}^d_i )^T, (\bm{0}_{p_{d+1}})^T,\ldots,
   (\bm{0}_{p_D})^T \Bigr).
 \end{equation}
Here, $\bm{0}_p\in \mathbb{R}^p$ is the vector with zero elements.  This
is a sparse coding \citep{olshausen2004sparse} in the sense that nonzero
elements for domains do not overlap each other.  All the $N$ vectors of
all domains are now represented as points in the same $\mathbb{R}^P$. We
will get the solution of the optimization problem of (\ref{eq:phia}) by
applying the single-domain version of the spectral graph embedding of
\cite{yan2007graph} to these $\bm{\tilde x}^d_i$ vectors.

In Section~\ref{sec:spectral}, we will review the spectral graph
embedding methods. In Section~\ref{sec:cdmca}, we will show that the
coding (\ref{eq:coding}) solves the minimization of (\ref{eq:phia}).  An
interesting connection to the classical associative memory model of
neural networks \citep{kohonen1972correlation,nakano1972associatron} is
also discussed there by noticing that coding $\bm{\tilde x}^d_i +
\bm{\tilde x}^e_j$ corresponds to the matching $w^{de}_{ij}$.  In
Section~\ref{sec:mcca}, the relations to the multivariate analysis
methods are explained. In Section~\ref{sec:example}, we show an
illustrative numerical example of cross-domain matching. In particular,
we discuss a cross-validation method for choosing the dimension $K$ of
the common space and a regularization parameter; we resample the
matching weights $w^{de}_{ij}$ instead of data vectors $\bm{x}^d_i$
there.

\section{A brief review of the spectral graph embedding} \label{sec:spectral}

\subsection{The spectral graph theory} \label{sec:spectralgraph}

Before discussing the cross-domain matching, here we review the spectral
graph theory \citep{chung1997spectral}. We then consider extra
constraints in Section~\ref{sec:graphembedding}. This result will be
used for the cross-domain matching in Section~\ref{sec:cdmca}. The
following argument is based on the spectral clustering, in particular
the normalized graph Laplacian \citep{shi2000normalized, ng2002spectral,
von2007tutorial} and the spectral embedding \citep{belkin2003laplacian}.

Let $N>0$ be the number of vertices of a graph, and these vertices are
represented by vectors $\bm{y}_i\in \mathbb{R}^K$, $i=1,\ldots,N$ of
dimension $K\le N$.  The weighted adjacency matrix is $\bm{W} = (w_{ij})
\in \mathbb{R}^{N\times N}$ with symmetric weights $w_{ij}=w_{ji}\ge
0$, $i,j=1,\ldots,N$. Let
$\bm{M}=\diag(\bm{W}\bm{1}_N)\in\mathbb{R}^{N\times N}$ be the diagonal
matrix with elements $\sum_{j=1}^N w_{ij}$, $i=1,\ldots,N$.  Here
$\bm{1}_N\in \mathbb{R}^N$ denotes the vector with all elements being 1.
The graph Laplacian is $\bm{M}-\bm{W}$.

For a given $\bm{W}$, we would like to find $\bm{y}_1,\ldots,\bm{y}_N$
that minimize the error function
\begin{equation} \label{eq:phiyn}
\phi(\bm{y}_1,\ldots,\bm{y}_N) = 
\frac{1}{2} \sum_{i=1}^N \sum_{j=1}^N w_{ij} \|\bm{y}_i - \bm{y}_j \|^2
\end{equation}
subject to certain constraints. For avoiding the trivial solution of all
zero vectors, we assume the constraints
\begin{equation} \label{eq:ymy}
 \bm{Y}^T \bm{M} \bm{Y} = \bm{I}_K,
\end{equation}
where $\bm{Y}\in \mathbb{R}^{N \times K}$ is defined by $\bm{Y}^T =
(\bm{y}_1,\ldots,\bm{y}_N)$, and $\bm{I}_K\in\mathbb{R}^{K\times K}$ is
the identity matrix. By simple rearrangement of the formula, we get
\[
 \trace(\bm{Y}^T \bm{W} \bm{Y}) = \sum_{i=1}^N \sum_{j=1}^N w_{ij}
      \bm{y}_i^T \bm{y}_j,\quad
 \trace(\bm{Y}^T \bm{M} \bm{Y}) = \sum_{i=1}^N \Bigl(\sum_{j=1}^N
      w_{ij}\Bigr)     \|\bm{y}_i\|^2.
\]
Thus the error function is rewritten as
\[
 \begin{split}
\phi(\bm{y}_1,\ldots,\bm{y}_N)   &= 
\frac{1}{2} \sum_{i=1}^N \sum_{j=1}^N w_{ij} 
\Bigl(\|\bm{y}_i\|^2 + \|\bm{y}_j \|^2 - 2 \bm{y}_i^T \bm{y}_j 
\Bigr)\\
&= \trace(\bm{Y}^T (\bm{M}-\bm{W}) \bm{Y}) \\
&= K - \trace(\bm{Y}^T \bm{W} \bm{Y}).
 \end{split}
\]
Therefore, minimization of (\ref{eq:phiyn}) is equivalent to
maximization of $ \trace(\bm{Y}^T \bm{W} \bm{Y})$.

Let $\bm{M}^{-1/2}\in \mathbb{R}^{N \times N}$ be the diagonal matrix
with elements $(\sum_{j=1}^N w_{ij})^{-1/2}$, $i=1,\ldots,N$. The
eigenvalues of $\bm{M}^{-1/2} \bm{W} \bm{M}^{-1/2}$ are $\lambda_1\ge
\lambda_2 \ge \cdots \ge \lambda_N$ and the corresponding normalized
eigenvectors are $\bm{u}_1,\ldots,\bm{u}_N\in \mathbb{R}^N$. The
solution of minimizing (\ref{eq:phiyn}) subject to (\ref{eq:ymy}) is
given by $\bm{Y} = \bm{M}^{-1/2} (\bm{u}_1,\ldots,\bm{u}_K)$.

\subsection{The spectral graph embedding for dimensionality reduction} \label{sec:graphembedding}

In addition to the constraints (\ref{eq:ymy}), \cite{yan2007graph}
introduced extra constraints that the column vectors of $\bm{Y}$ are
included in a specified linear subspace.  Let us specify $\bm{x}_i \in
\mathbb{R}^P$, $i=1,\ldots,N$ with some $K \le P\le N$. Define the data
matrix $\bm{X}\in\mathbb{R}^{N\times P}$ by
$\bm{X}^T=(\bm{x}_1,\ldots,\bm{x}_N)$. We assume that $\bm{Y}$ is
expressed in the form
\begin{equation} \label{eq:yxa}
 \bm{Y} = \bm{X} \bm{A}
\end{equation}
using an arbitrary matrix $\bm{A}\in \mathbb{R}^{P\times K}$.  Therefore,
minimization of (\ref{eq:phiyn}) is equivalent to finding $\bm{A}$ that
maximizes $ \trace(\bm{A}^T \bm{X}^T \bm{W} \bm{X} \bm{A})$ subject to
the constraints $\bm{A}^T \bm{X}^T \bm{M} \bm{X} \bm{A} = \bm{I}_K$.

For numerical stability, we introduce quadratic regularization terms
similar to those of \cite{takane2008regularized}. First, we define
two $P\times P$ matrices by
\[
 \bm{G} = \bm{X}^T \bm{M} \bm{X} + \gamma_M \bm{L}_M,\quad
 \bm{H} = \bm{X}^T \bm{W} \bm{X} + \gamma_W \bm{L}_W.
\]
Here $\gamma_M, \gamma_W \in \mathbb{R}$ are regularization parameters,
and $\bm{L}_M, \bm{L}_W\in\mathbb{R}^{P \times P}$
are non-negative definite, typically  $\bm{L}_M=\bm{L}_W=\bm{I}_P$.
Then, we consider the optimization problem:
\begin{gather}
\mbox{Maximize}\quad \trace( \bm{A}^T \bm{H} \bm{A} )\quad\mbox{with
 respect to}\quad\bm{A}\in\mathbb{R}^{P\times K} \label{eq:aha}\\
\mbox{subject to}\quad \bm{A}^T \bm{G} \bm{A} = \bm{I}_K \label{eq:aga}.
\end{gather}       
This reduces to the problem of Section~\ref{sec:spectralgraph} by
letting $\bm{X}=\bm{I}_N$, $\gamma_M=\gamma_W=0$.  For the solution of
the optimization problem, we denote $\bm{G}^{1/2}\in \mathbb{R}^{P\times
P}$ be one of the matrices satisfying $(\bm{G}^{1/2})^T \bm{G}^{1/2} =
\bm{G}$. The inverse matrix is denoted by $\bm{G}^{-1/2} = (\bm{G}^{1/2}
)^{-1}$. These are easily computed by, say, Cholesky decomposition or
spectral decomposition of symmetric matrix.  The eigenvalues of
$(\bm{G}^{-1/2})^T \bm{H} \bm{G}^{-1/2}$ are $\lambda_1\ge\lambda_2\ge \cdots
 \ge \lambda_P$, and the corresponding normalized eigenvectors are
$\bm{u}_1,\bm{u}_2,\ldots,\bm{u}_P \in \mathbb{R}^P$.
The solution of our optimization problem is
\begin{equation} \label{eq:agu}
 \bm{A} = \bm{G}^{-1/2}(\bm{u}_1,\ldots,\bm{u}_K).
\end{equation}       

To see what we are actually solving, let us rewrite the error function
(\ref{eq:phiyn}) with respect to $\bm{A}$ under the constraints
(\ref{eq:yxa}) and (\ref{eq:aga}).
\[
 \begin{split}
  \phi(\bm{A}) 
& = \trace(\bm{Y}^T (\bm{M}-\bm{W}) \bm{Y}) \\
& = \trace(\bm{A}^T \bm{X}^T (\bm{M}-\bm{W}) \bm{X}\bm{A} )\\
  & = \trace( \bm{A}^T ( \bm{G} - \bm{H} - \gamma_M \bm{L}_M + \gamma_W
  \bm{L}_W) \bm{A})\\
  & = K - \trace(\bm{A}^T \bm{H} \bm{A}) 
  - \trace(\bm{A}^T ( \gamma_M \bm{L}_M - \gamma_W  \bm{L}_W) \bm{A})
 \end{split}
\]
Thus, maximization of (\ref{eq:aha}) subject to (\ref{eq:aga})
is equivalent to  minimization of
\begin{equation}\label{eq:phiagamma}
  \phi(\bm{A}) + \trace(\bm{A}^T ( \gamma_M \bm{L}_M - \gamma_W  \bm{L}_W) \bm{A}).
\end{equation}
For the second term working as a regularization term properly, $\gamma_M
       \bm{L}_M - \gamma_W \bm{L}_W$ should be nonnegative definite.

\section{Cross-domain matching correlation analysis} \label{sec:cdmca}

Now we are back to the cross-domain matching. We define several matrices
for rewriting (\ref{eq:yax}) and (\ref{eq:phia}) in a simple form.  The
data matrices $\bm{X}^d \in \mathbb{R}^{n_d \times p_d}$ for domains
$d=1,\ldots,D$ are defined by $( \bm{X}^d ) ^T =
(\bm{x}^d_1,\ldots,\bm{x}^d_{n_d})$.  We put these $D$ matrices in the
block diagonal positions of a $N \times P$ matrix to define a large data
matrix $\bm{X}=\Diag(\bm{X}^1,\ldots,\bm{X}^D) \in \mathbb{R}^{N \times P}$.
We concatenate the transformation matrices to define $\bm{A}\in
\mathbb{R}^{P \times K}$ as $ \bm{A}^T = ((\bm{A}^1)^T,\ldots,
(\bm{A}^D)^T)$. The vectors in the common space are also concatenated
to define $\bm{Y}^d\in \mathbb{R}^{n_d \times K}$ and $\bm{Y}\in
\mathbb{R}^{N \times K}$ as $( \bm{Y}^d )^T = (\bm{y}^d_1,
\ldots,\bm{y}^d_{n_d})$, $\bm{Y}^T = ((\bm{Y}^1)^T,\ldots,
(\bm{Y}^D)^T)$. The matching weight matrices are $\bm{W}^{de} =
(w^{de}_{ij}) \in \mathbb{R}^{n_d \times n_e}$ for $d,e=1,\ldots,D$, and
they are placed in a array to define $\bm{W} = (\bm{W}^{de}) \in
\mathbb{R}^{ N \times N}$.

Using these matrices, the transformation (\ref{eq:yax}) is written as
(\ref{eq:yxa}), and the error function (\ref{eq:phia}) is written as
(\ref{eq:phiyn}) or $\phi(\bm{A})$ of Section~\ref{sec:graphembedding}.
Adding the regularization term to the error function, the objective
function becomes (\ref{eq:phiagamma}), and the solution is
(\ref{eq:agu}). Thus, the cross-domain matching is solved by the
single-domain version of the spectral graph embedding. An important
point is that the large data matrix $\bm{X}$ is expressed as
\[
 \bm{X}^T = (\bm{\tilde x}^1_1,\ldots,\bm{\tilde x}^1_{n_1},\ldots,
\bm{\tilde x}^D_1,\ldots,\bm{\tilde x}^D_{n_D}),
\]
meaning $\bm{X}$ is the data matrix consists of the augmented
vectors. What we have done is, therefore, interpreted as simply applying
the spectral graph embedding of \cite{yan2007graph} to the $N$
augmented vectors in $\mathbb{R}^P$.

It would be better to rewrite the constraints (\ref{eq:aga}) in terms of
$\bm{A}^1,\ldots,\bm{A}^D$ for cross-domain matching.  Notice
$\bm{M}=\Diag(\bm{M}^1,\ldots,\bm{M}^D)$ with $\bm{M}^d = \diag(
(\bm{W}^{d1},\ldots,\bm{W}^{dD})\bm{1}_N)$, and so $\bm{X}^T \bm{M}
\bm{X} = \Diag((\bm{X}^1)^T \bm{M}^1 \bm{X}^1,\ldots, (\bm{X}^D)^T
\bm{M}^D \bm{X}^D )$. For simplicity, we assume that the regularization
matrix is written as a block diagonal matrix as
$\bm{L}_M=\Diag(\bm{L}_M^1,\ldots,\bm{L}_M^D)$. Then we have
\[
  \bm{A}^T \bm{G} \bm{A} = \sum_{d=1}^D (\bm{A}^d)^T \Bigl(
   (\bm{X}^d)^T \bm{M}^d \bm{X}^d + \gamma_M \bm{L}_M^d \Bigr)
   \bm{A}^d = \bm{I}_K.
\]
This is expressed for the vectors in $\bm{A}^d$ as
\begin{equation} \label{eq:xmxdelta}
   \sum_{d=1}^D (\bm{a}_k^d)^T  \Bigl( (\bm{X}^d)^T \bm{M}^d
       \bm{X}^d + \gamma_M \bm{L}_M^d \Bigr) \bm{a}_l^d =
       \delta_{kl},\quad
       k,l=1,\ldots,K
\end{equation}       
using the Kronecker delta.

As a final remark of this section, we discuss a coding of matching for
further implications. Let $E$ be the number of nonzero elements in the
lower triangular part of $\bm{W}$. In other words, $E$ is the number of
edges in the graph.  We define a diagonal matrix $\bm{\breve
W}\in\mathbb{R}^{E\times E}$ with elements of these nonzero
$\{w^{de}_{ij}\}$. Instead of working on the vertices of the graph, here
we work on the edges of the graph for data analysis. So, the data vector
is now coded as $\bm{\tilde x}^d_i + \bm{\tilde x}^e_j$ for the matching
weight $w^{de}_{ij}$. We define the data matrix $\bm{\breve X}\in
\mathbb{R}^{E \times P}$ by concatenating $\bm{\tilde x}^d_i +
\bm{\tilde x}^e_j$ in the same order as $\bm{\breve W}$.  Since $
\bm{\breve X}^T \bm{\breve W} \bm{\breve X} = \bm{X}^T \bm{M} \bm{X} +
\bm{X}^T \bm{W} \bm{X} $, minimization of (\ref{eq:phiagamma}) is
equivalent to maximization of $ \trace(\bm{A}^T ( \bm{\breve X}^T
\bm{\breve W} \bm{\breve X} + \gamma_M \bm{L}_M + \gamma_W \bm{L}_W
)\bm{A} )$. Therefore, the cross-domain matching is interpreted as a
kind of PCA for input patterns coded as $\bm{\tilde x}^d_i + \bm{\tilde
x}^e_j$. Interestingly, this idea is found in one of the classical
neural network models.  Any part of the memorized vector can be used as
a key for recalling the whole vector in the auto-associative correlation
matrix memory \citep{kohonen1972correlation,
nakano1972associatron}. This associative memory may recall $\bm{\tilde
x}^d_i + \bm{\tilde x}^e_j$ for input key either $\bm{\tilde x}^d_i$ or
$\bm{\tilde x}^e_j$. It would be a subject of future research to work on
$\bm{\tilde x}^{d}_i + \bm{\tilde x}^{e}_j + \bm{\tilde x}^{f}_k
+\cdots$ for joint associations of three or more vectors.

\section{Relations to multiset canonical correlation analysis} \label{sec:mcca}

In this section, we assume that the numbers of vectors are the same for
all domains. Then the cross-domain matching reduces to a classical
multivariate analysis of statistics. Let $n_1=\cdots=n_D=n$, and $N=nD$.
We assume that the weight matrix is specified as $\bm{W}^{de}=c_{de}
\bm{I}_n$ using a coefficient $c_{de}\ge0$ for all $d,e=1,\ldots,D$.  In
this case, the cross-domain matching becomes a version of MCCA, where
connections between sets of variables are specified by the coefficients
$c_{de}$ \citep{tenenhaus2011regularized}. Another version of MCCA with
all $c_{de}=1$ is discussed extensively in \cite{takane2008regularized}.

Here we show how the objective function (\ref{eq:aha}) and the
constraints (\ref{eq:aga}) are expressed in the case of MCCA.  Noting
that $\bm{X}^T \bm{W} \bm{X}$ is an array of $(\bm{X}^d)^T \bm{W}^{de}
\bm{X}^e =c_{de }(\bm{X}^d)^T \bm{X}^e $, $d,e=1,\ldots,D$, we have
\begin{equation} \label{eq:ahacde}
\begin{split}
\trace(\bm{A}^T \bm{H} \bm{A})
&=\sum_{d=1}^D \sum_{e=1}^D c_{de} \trace((\bm{A}^d)^T
(\bm{X}^d)^T \bm{X}^e \bm{A}^e) + \gamma_W \sum_{d=1}^D
 \trace((\bm{A}^d)^T \bm{L}_W^d \bm{A}^d)\\
 &=\sum_{k=1}^K \Bigl(
 \sum_{d=1}^D \sum_{e=1}^D c_{de} (\bm{a}_k^d)^T (\bm{X}^d)^T
 \bm{X}^e \bm{a}_k^e
 +\sum_{d=1}^D \gamma_W (\bm{a}_k^d)^T \bm{L}_W^d \bm{a}_k^d
 \Bigr)
\end{split}
\end{equation}
For simplicity, we assumed that the regularization matrix is written as
a block diagonal matrix as $\bm{L}_W =
\Diag(\bm{L}_W^1,\ldots,\bm{L}_W^D)$.  The constraints (\ref{eq:aga})
are expressed as (\ref{eq:xmxdelta}) with $\bm{M}^d = (\sum_{e=1}^D
c_{de}) \bm{I}_n$. The constraints correspond to eq.~(31) of
\cite{takane2008regularized} except for a difference in scaling, when
$c_{de}=1$, $\bm{L}_M=\bm{L}_W$, and $\gamma_M = D \gamma_W$.

Further assume that $p_1=\cdots=p_D=1$ and $P=D$. Each $\bm{X}^d\in
\mathbb{R}^{n \times 1} $ is a vector now.  $(\bm{G}^{-1/2})^T \bm{H}
\bm{G}^{-1/2}$ becomes the sample correlation matrix scaled by the
factor $D^{-1}$. Thus, the cross-domain matching is equivalent to PCA.

 \section{An illustrative numerical example} \label{sec:example}

 \subsection{Data generation} \label{sec:data}

 We look at a very simple example to see how the methods work.  We
 randomly generated a data with $D=3$, $p_1=10, p_2=30, p_3=100$,
 $n_1=125, n_2=250, n_3=500$ in the following steps.
\begin{itemize}
 \item[1.] We placed
 points on $5 \times 5$ grid in $\mathbb{R}^2$ as
 $(1,1),(1,2),(1,3),(1,4),(1,5), (2,1),\ldots,(5,5)$.  They are
 $(\bm{x}^0_1)^T,\ldots,(\bm{x}^0_{25})^T$, where $d=0$ is treated as a
 special domain for data generation. These 25 values are repeatedly used
	   to define $\bm{x}^0_i$ for $i=26,27,\ldots$.
 \item[2.] We made random matrices $\bm{B}^d \in \mathbb{R}^{p_d \times
	   2}$, $d=1,2,3$, with all elements distributed as $N(0,1)$
	   independently. Then, we generated data vectors
	   $\bm{x}_i^d=\bm{B}^d \bm{x}^0_i + \bm{\epsilon}^d_i$,
	   $i=1,\ldots,n_d$. Elements of $\bm{\epsilon}^d_i$ are
	   distributed as $N(0,0.5^2)$ independently.
	   Each column of $\bm{X}^d$ is standardized to mean zero and
	   variance one.
 \item[3.] The numbers of data vectors $\bm{x}^d_i$ generated from each
	   grid point are 5, 10, 20, respectively, for $d=1,2,3$. For
	   defining underlying true associations, we linked these 35
	   vectors to each other, except for those within a same domain.
	   The true weights for these 35 vectors are $\bar
	   w^{de}_{ij}=1$ for $d\neq e$, $\bar w^{dd}_{ij}=0$. All other
	   weights across grid points are zero.  The numbers of nonzero
	   elements (lower triangular) are 1250, 2500, 5000 (total
	   8750), respectively, for $\bm{\bar W}^{21}$, $\bm{\bar
	   W}^{31}$, $\bm{\bar W}^{32}$.
 \item[4.] We made weight matrices $\bm{W}^{de}$ by randomly sampling
	   2\% of links from $\bm{\bar W}^{de}$. The numbers of nonzero
	   elements (lower triangular) became 28, 50, 97 (total 175),
	   respectively, for $\bm{W}^{21}, \bm{W}^{31}, \bm{W}^{32}$.
\end{itemize}

\subsection{Finding the common space} \label{sec:analysis}

We applied the cross-domain matching with $\gamma_W=0$, $\gamma_M=0.1$
to the generated data . The regularization matrix is block diagonal
$\bm{L}_M =\Diag(\bm{L}_M^1,\bm{L}_M^2,\bm{L}_M^3)$ with $\bm{L}_M^d =
\alpha_d \bm{I}_{p_d}$ and $\alpha_d=\trace((\bm{X}^d)^T \bm{M}^d
\bm{X}^d)/p_d$. The results are shown in Fig.~\ref{fig:fig1} and
Fig.~\ref{fig:fig2}.

Like PCA, we denote PC$k$ for the $k$-th component of the common space
$(\bm{y}_i^d)_k$. Scatter plots of data vectors in the common space are
shown in Fig.~\ref{fig:pc12} and Fig.~\ref{fig:pc13}.  The $5 \times 5$
structure is clearly observed in (PC1, PC2), while PC3 looks almost
noise. Here each PC is rescaled to have unweighted variance 1.  Looking
at eigenvalues $\lambda_k$ (correspond to the canonical correlations of
CCA) in Fig.~\ref{fig:cor}, they are almost 1 for PC1 and PC2, and
decrease rapidly for $k\ge 3$, indicating $K=2$ is a good choice. The
number of positive $\lambda_k$ is 40 ($=p_1+p_2$ in this example). We only
look at these 40 PC's, because negative $\lambda_k$ are due to change of
the sign of axes.

Picking a data vector ($d=2, i=1$) as a query, and look for vectors
close to it in the common space. This query vector can be treated as a
new input, because it was not linked to any other vectors in
$\bm{W}^{de}$.  In Fig.~\ref{fig:distances}, distances to other vectors
$\|\bm{y}_i^d - \bm{y}_1^2\|$, $i=1,\ldots,n_d$, $d=1,\ldots,D$ are
computed with $K=2$. The ``true'' distances are computed as
$\|\bm{x}^0_i - \bm{x}^0_1 \|$. They agree very well, meaning that we
will find closely related vectors.

What happens if we use a wrong $K$? Results are shown in
Fig.~\ref{fig:fig2}. The observed distances in the common space are
disturbed by PC3 in Fig.~\ref{fig:distances3}. The situation becomes
worse in Fig.~\ref{fig:distances20}, and it is not possible to make a
reasonable data-retrieval any more.  It is very important to choose an
appropriate $K$.

 \subsection{Choosing $K$ and $\gamma_M$ by cross-validation}
 \label{sec:crossvalidation}

 We write $\bm{A}=\bm{A}(\bm{W},\gamma_M)$ for (\ref{eq:agu}) and
 $\phi(\bm{A},\bm{W})$ for (\ref{eq:phia}) by omitting $\bm{X}$ from the
 notation. The error function is decomposed into each PC$k$ as
$\phi(\bm{A},\bm{W})=\sum_{k=1}^K \phi_k(\bm{A},\bm{W})$ with 
\[
       \phi_k(\bm{A},\bm{W}) = \frac{1}{2}\sum_{d=1}^D \sum_{e=1}^D
       \sum_{i=1}^{n_d} \sum_{j=1}^{n_e}
       w^{de}_{ij} \Bigl((\bm{y}_i^d)_k - (\bm{y}_j^e)_k  \Bigr)^2.
\]
The weights $\bm{W}=(w_{ij})$ are always rescaled to have $\sum_{i=1}^N
\sum_{j=1}^N w_{ij}=1$ in the computation of $\phi_k(\cdot, \bm{W})$
below.  For verifying an appropriate value for $K$ and $\gamma_M$, the
 error $\phi_k(\bm{A}(\bm{W},\gamma_M),\bm{\bar W})$ with respect to
the true weights $\bar w^{de}_{ij}$ is computed for $\gamma_M=0, 0.001,
0.01, 0.1, 1.0$ in Fig.~\ref{fig:trueerror}. The error is small for
$k=1,2$, and it rapidly increases for $k\ge 3$, confirming $K=2$ is the
right choice. Also, we confirm that the errors in PC1 and PC2 are
minimized when $\gamma_M=0.1$.

The values of $\bar w^{de}_{ij}$ are unknown in reality, and we have to
compute the error only from the observed $w^{de}_{ij}$.  However, the
fitting error $\phi_k(\bm{A}(\bm{W},\gamma_M),\bm{W})$ in
Fig.~\ref{fig:fittingerror} does not work well. The fitting error is
minimized when $\gamma_M=0$, but prediction of unlinked pairs of vectors
is not good as seen in Fig.~\ref{fig:g0pc12}. Another issue we notice in
Fig.~\ref{fig:fittingerror} is that the fitting error for $\gamma_M=0$
is not monotone increasing in PC; this becomes monotone when we rescale
 PC$k$ by factor $( \sum_{i} (\sum_j w_{ij}) (\bm{y}_i)_k / \sum_i
 \sum_j w_{ij})^{-1/2}$.

For estimating the true error, we then performed cross-validation
analysis as follows.  10\% of nonzero elements (lower triangular) of
$\bm{W}$ are resampled to make $\bm{W}^*$. In other words, the elements
of $\bm{W}^*$ are defined as $w^{*de}_{ij} = w^{de}_{ij} z^{*de}_{ij}$;
$z^{*de}_{ij}$ are generated by the Bernoulli trial with $P(z^{*de}_{ij}
=1 )=0.1$ and $P(z^{de}_{ij} =0 )=0.9$.  The number of nonzero
elements (lower triangular) of $\bm{W}^*$ was 19, and that of the
remaining matrix $\bm{W}-\bm{W}^*$ was 156, from which we computed
$\phi_k(\bm{A}((\bm{W}-\bm{W}^*)/0.9,\gamma_M),\bm{W}^*)$.  By repeating this
process 30 times, we computed the average error.  This cross-validation
error is shown in Fig.~\ref{fig:crossvalidation}. The plot is very
similar to Fig.~\ref{fig:trueerror}, and we successfully choose $K=2$,
$\gamma_M=0.1$. 
In fact, \cite{shimodaira2015crossvalidation} showed asymptotically as $N\to\infty$ that the 
cross-validation error unbiasedly estimates the true error by adjusting the bias of the fitting error.

\section*{Acknowledgments}

I would like to thank Kazuki Fukui and Haruhisa Nagata for helpful
discussions.

\bibliographystyle{imsart-nameyear}
\bibliography{stat2014}

\clearpage
\newlength{\figwidth}
\setlength{\figwidth}{6.5cm}

\begin{figure}
 \begin{center}
  \subfigure[(PC1, PC2)]{
  \includegraphics[width=\figwidth,clip]{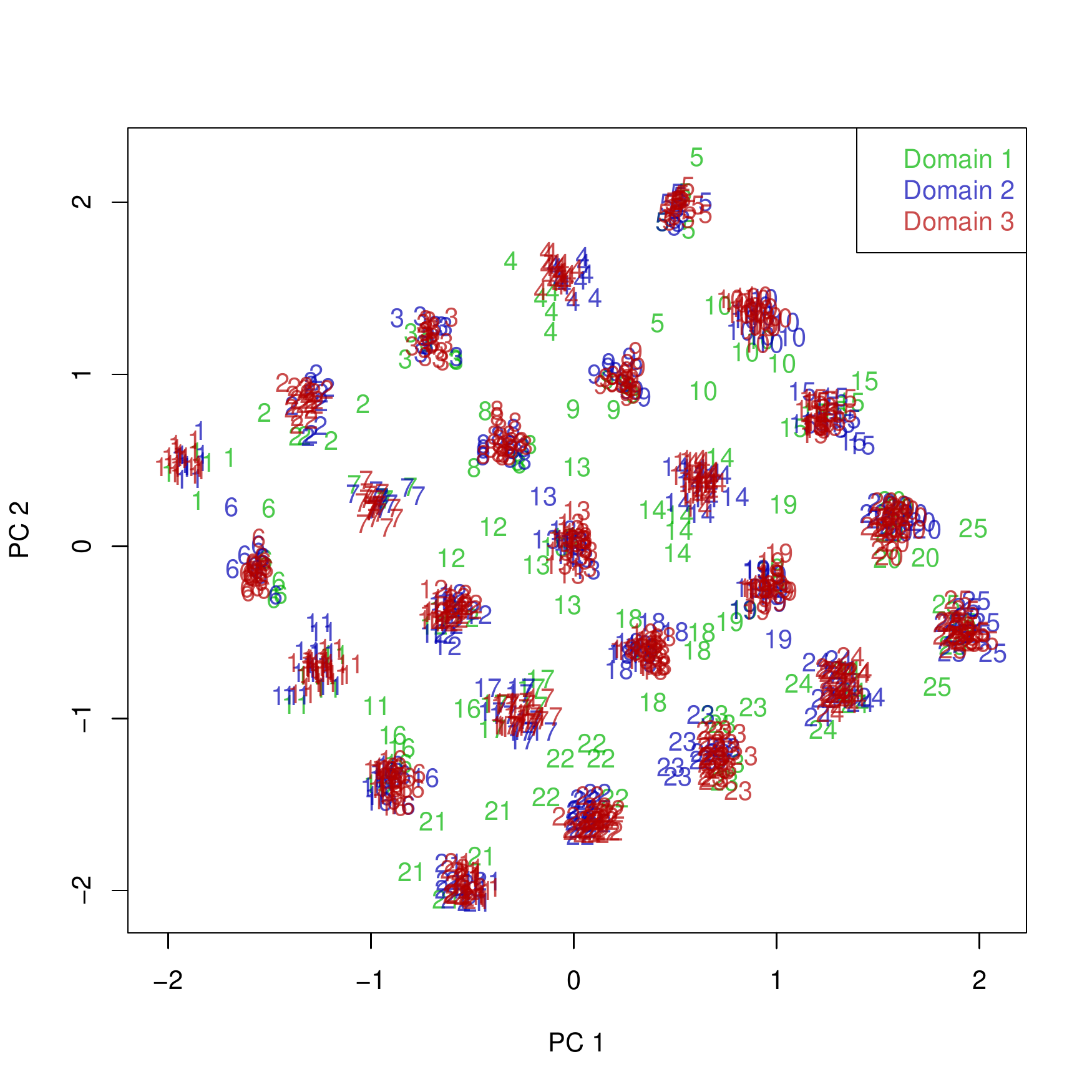}
  \label{fig:pc12}
  }
  \subfigure[(PC1, PC3)]{
  \includegraphics[width=\figwidth,clip]{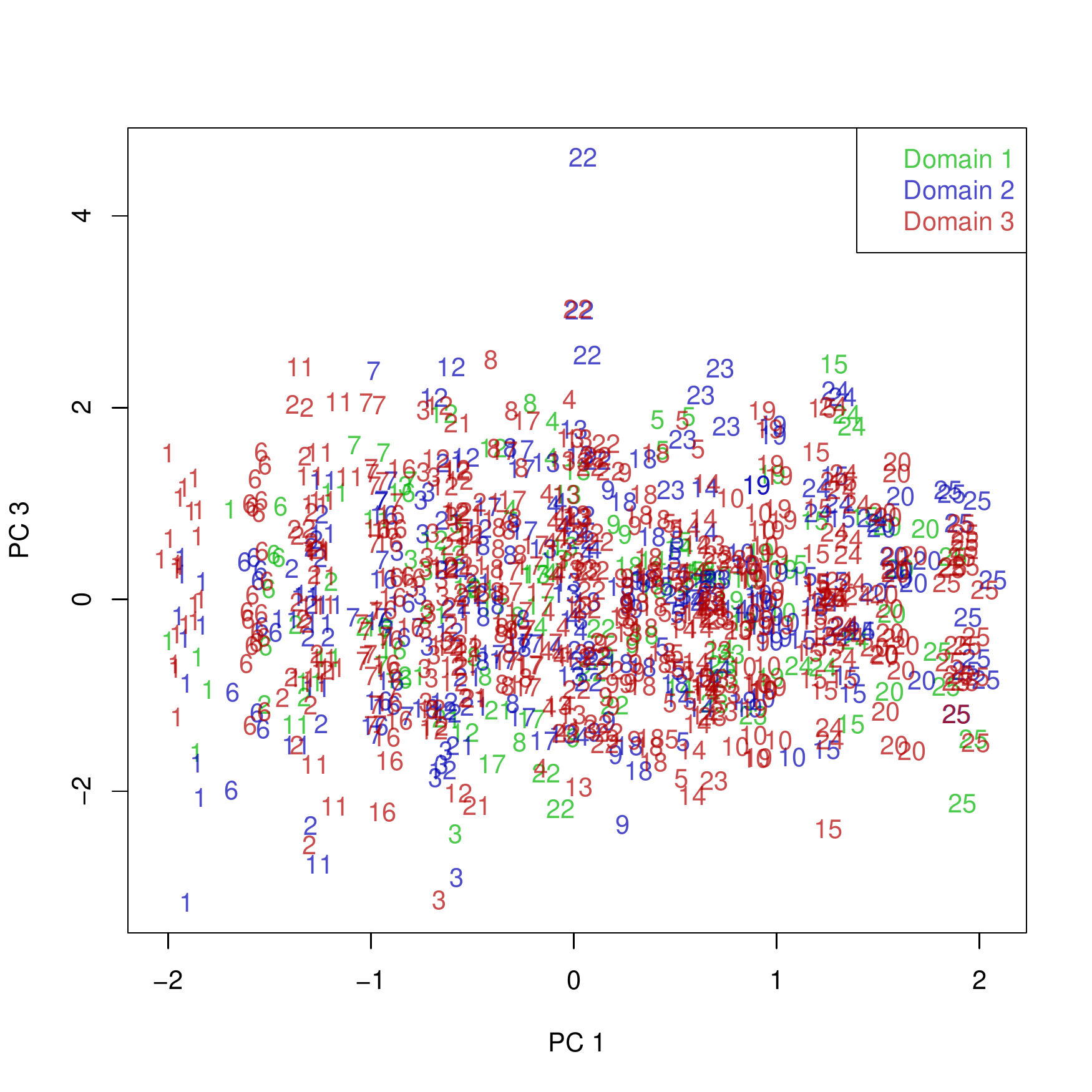}
  \label{fig:pc13}
  }\\
  \subfigure[Eigenvalues]{
  \includegraphics[width=\figwidth,clip]{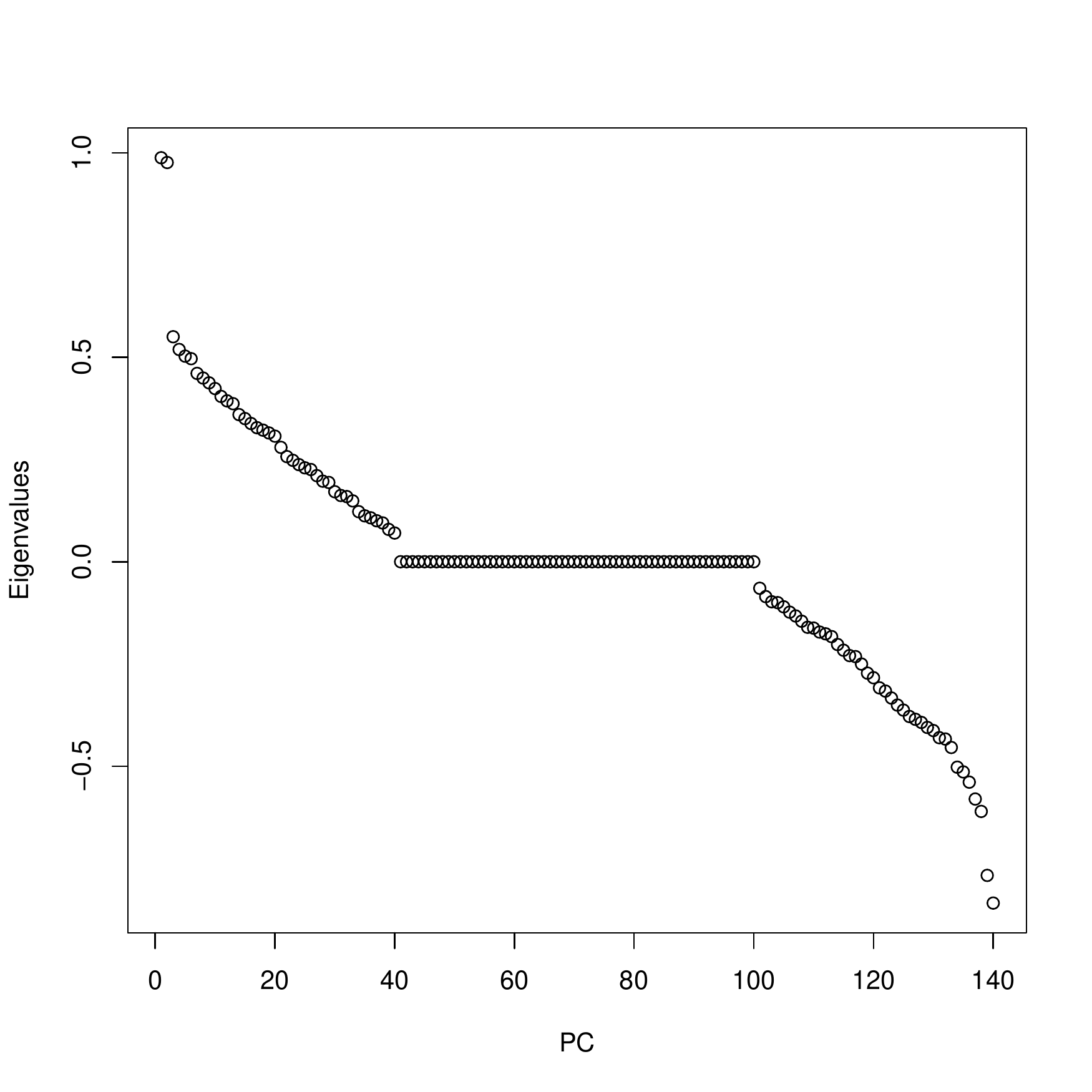}
  \label{fig:cor}
  }
  \subfigure[Distances from $\bm{y}^2_1$ ($K=2$)]{
  \includegraphics[width=\figwidth,clip]{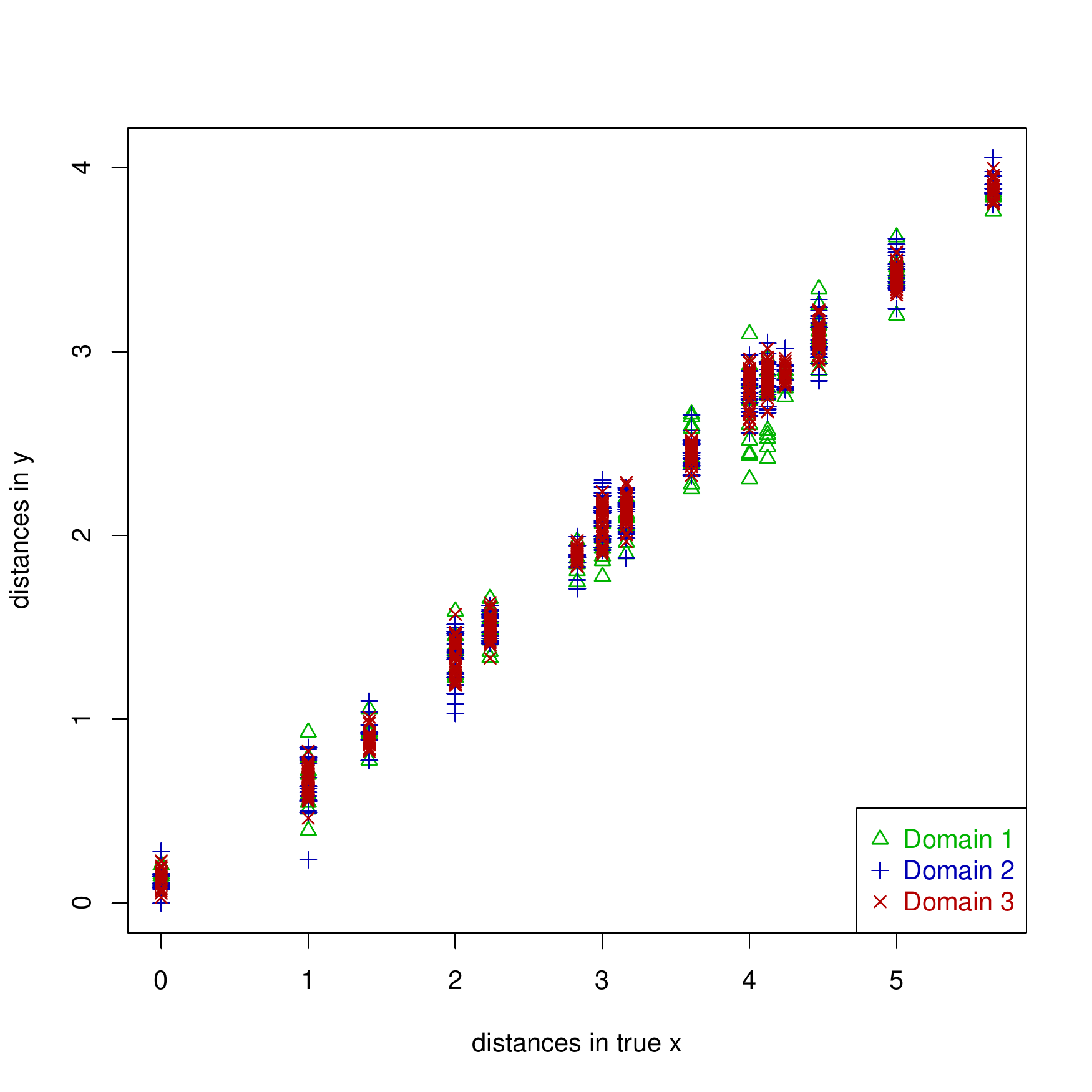}
  \label{fig:distances}
  }
 \end{center}
 \caption{Cross-domain matching of Section~\ref{sec:analysis} ($\gamma_M
 = 0.1$)} \label{fig:fig1}
\end{figure}

\begin{figure}
 \begin{center}
  \subfigure[Distances from $\bm{y}^2_1$ ($K=3$)]{
  \includegraphics[width=\figwidth,clip]{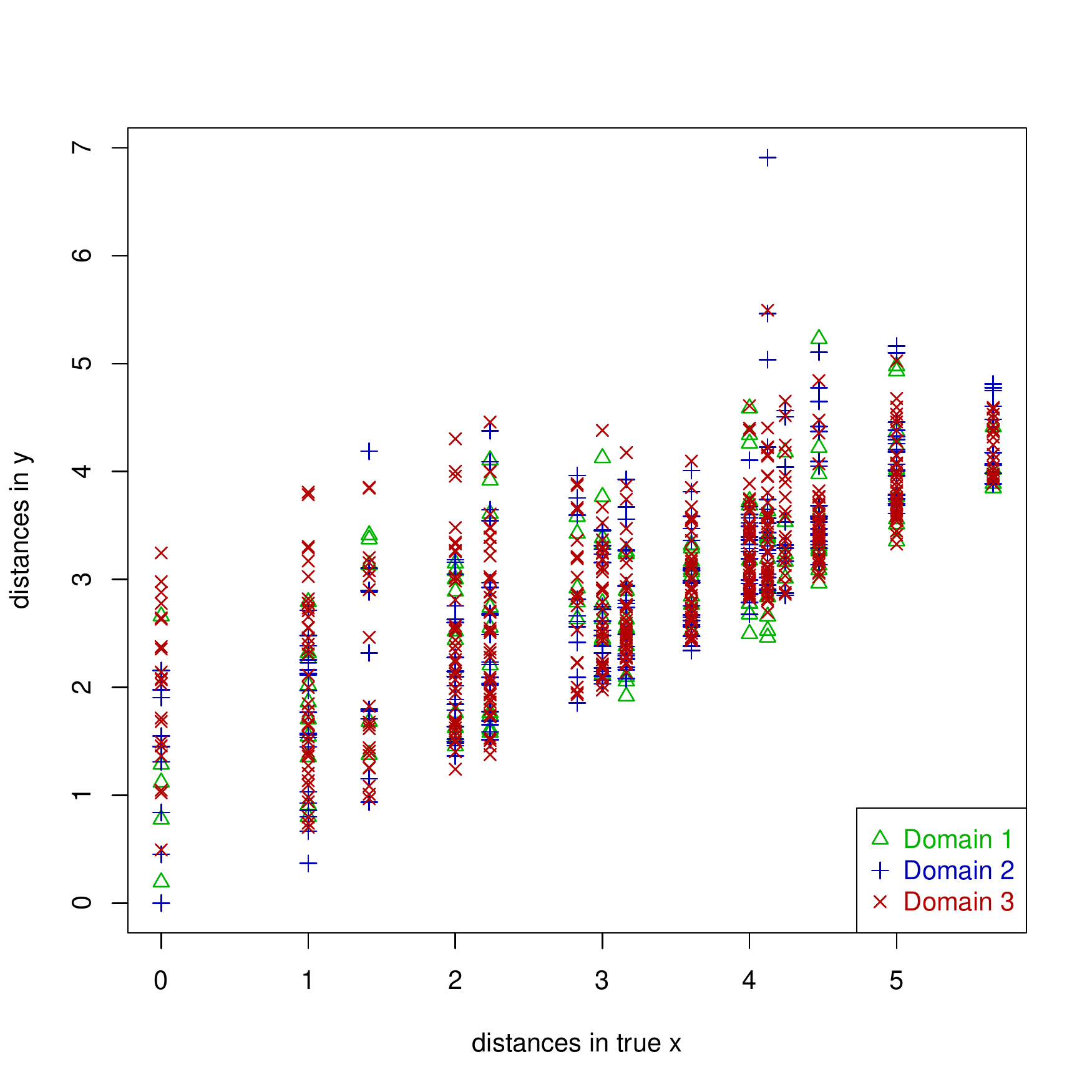}
  \label{fig:distances3}
  }
  \subfigure[Distances from $\bm{y}^2_1$ ($K=20$)]{
  \includegraphics[width=\figwidth,clip]{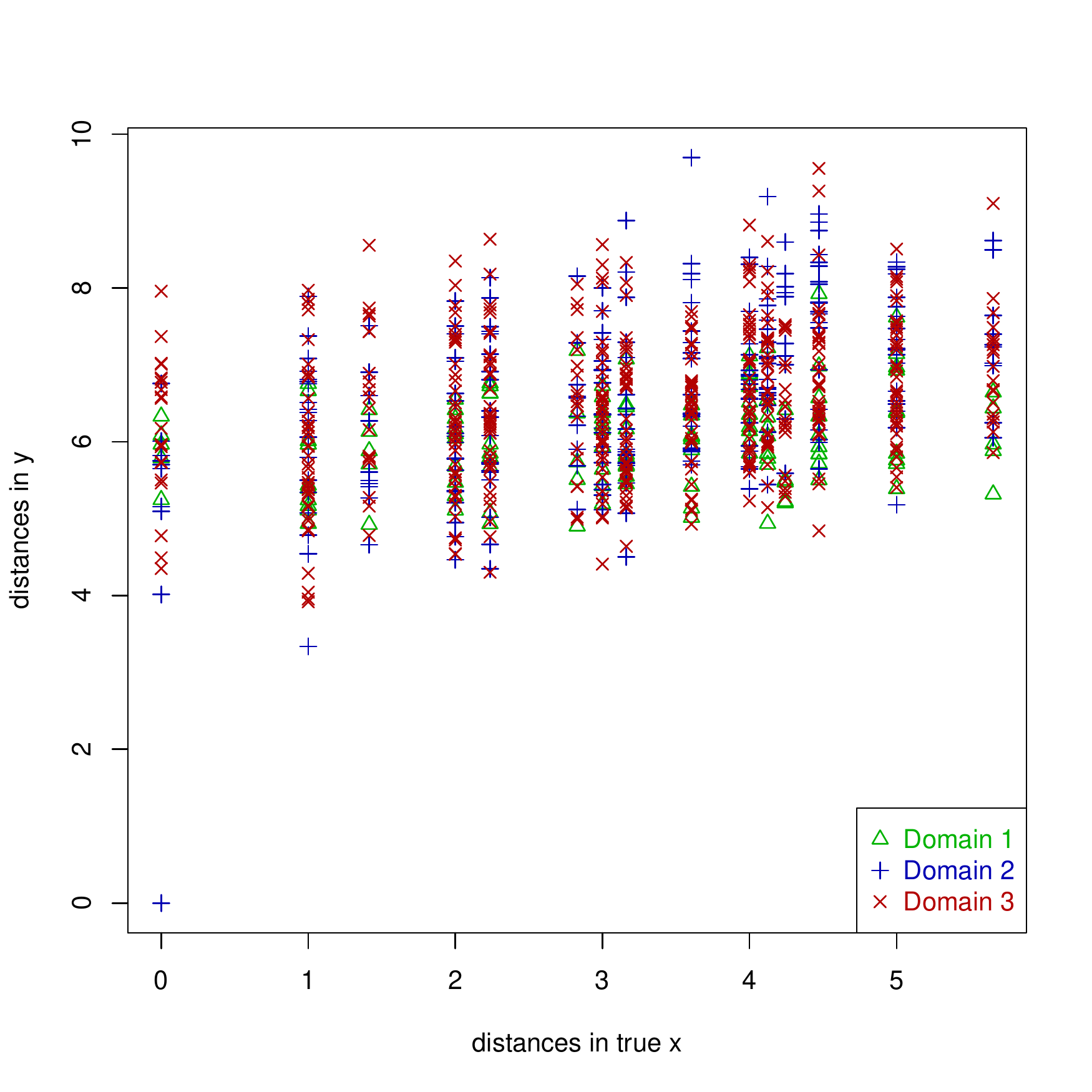}
  \label{fig:distances20}
  }
 \end{center}
 \caption{Using larger $K$  ($\gamma_M = 0.1$)}   \label{fig:fig2}
\end{figure}

\begin{figure}
  \begin{center}
  \subfigure[True error of each PC]{
  \includegraphics[width=\figwidth,clip]{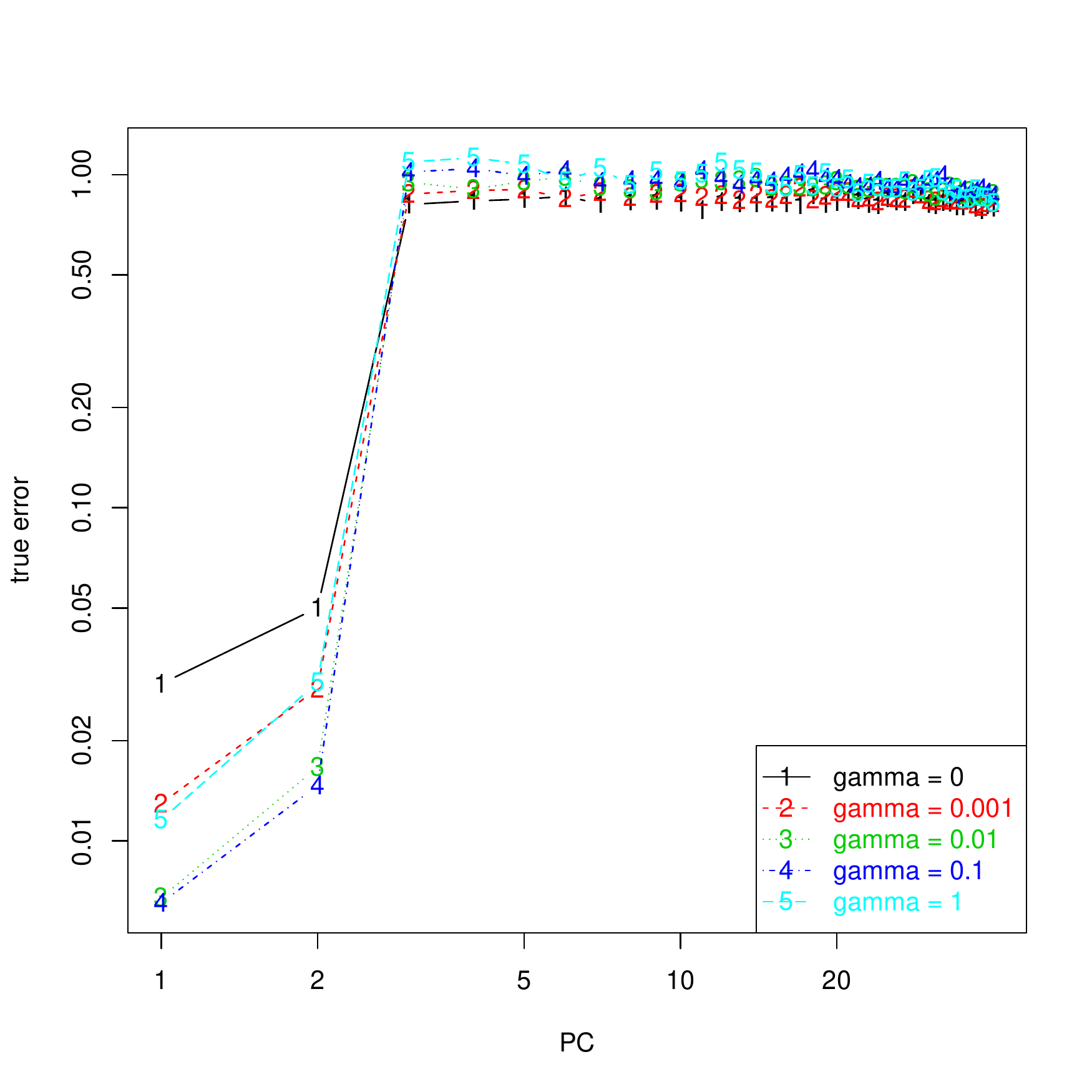}
  \label{fig:trueerror}
   }
  \subfigure[Fitting error of each PC]{
  \includegraphics[width=\figwidth,clip]{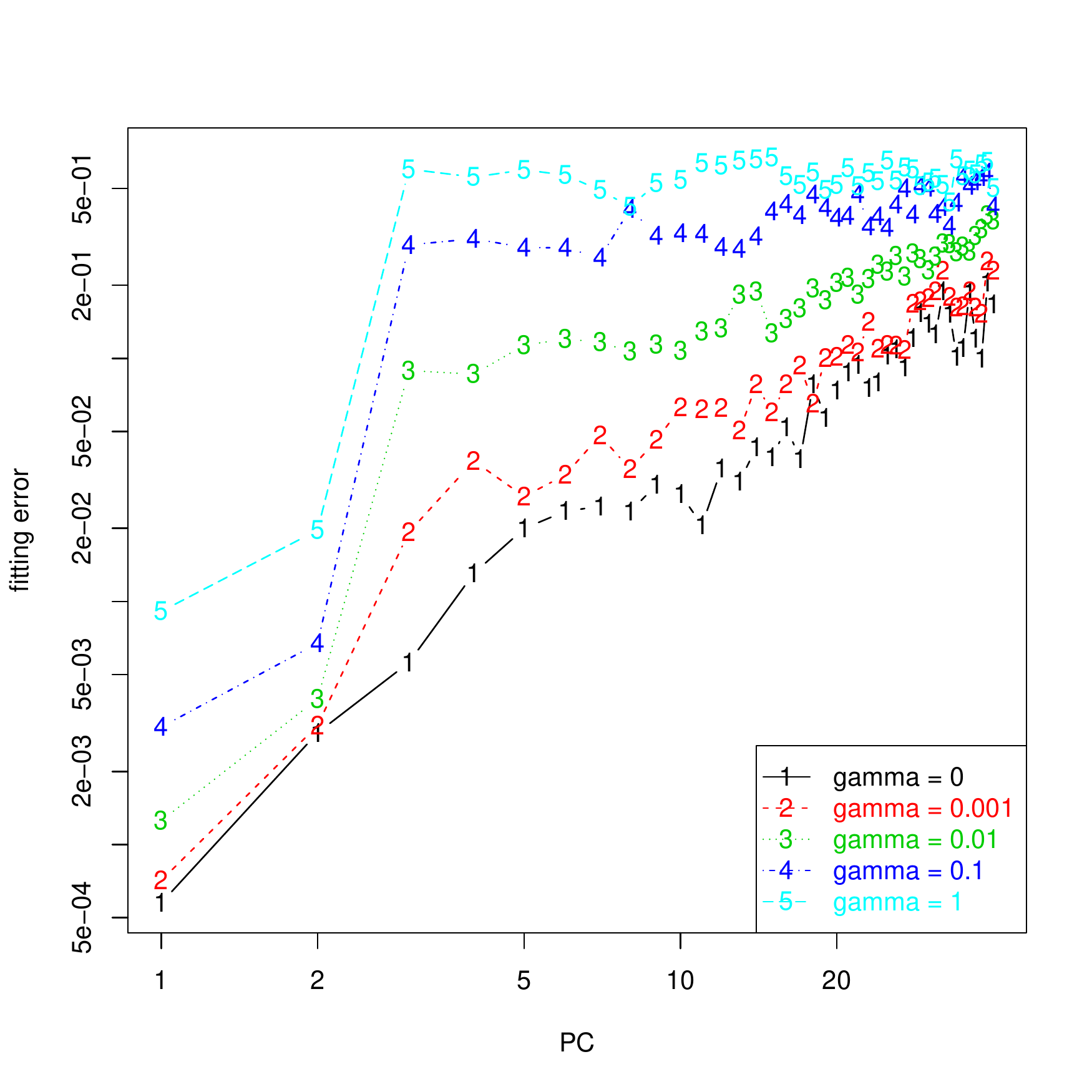}
  \label{fig:fittingerror}
   }\\
  \subfigure[(PC1, PC2) with $\gamma_M=0$]{
  \includegraphics[width=\figwidth,clip]{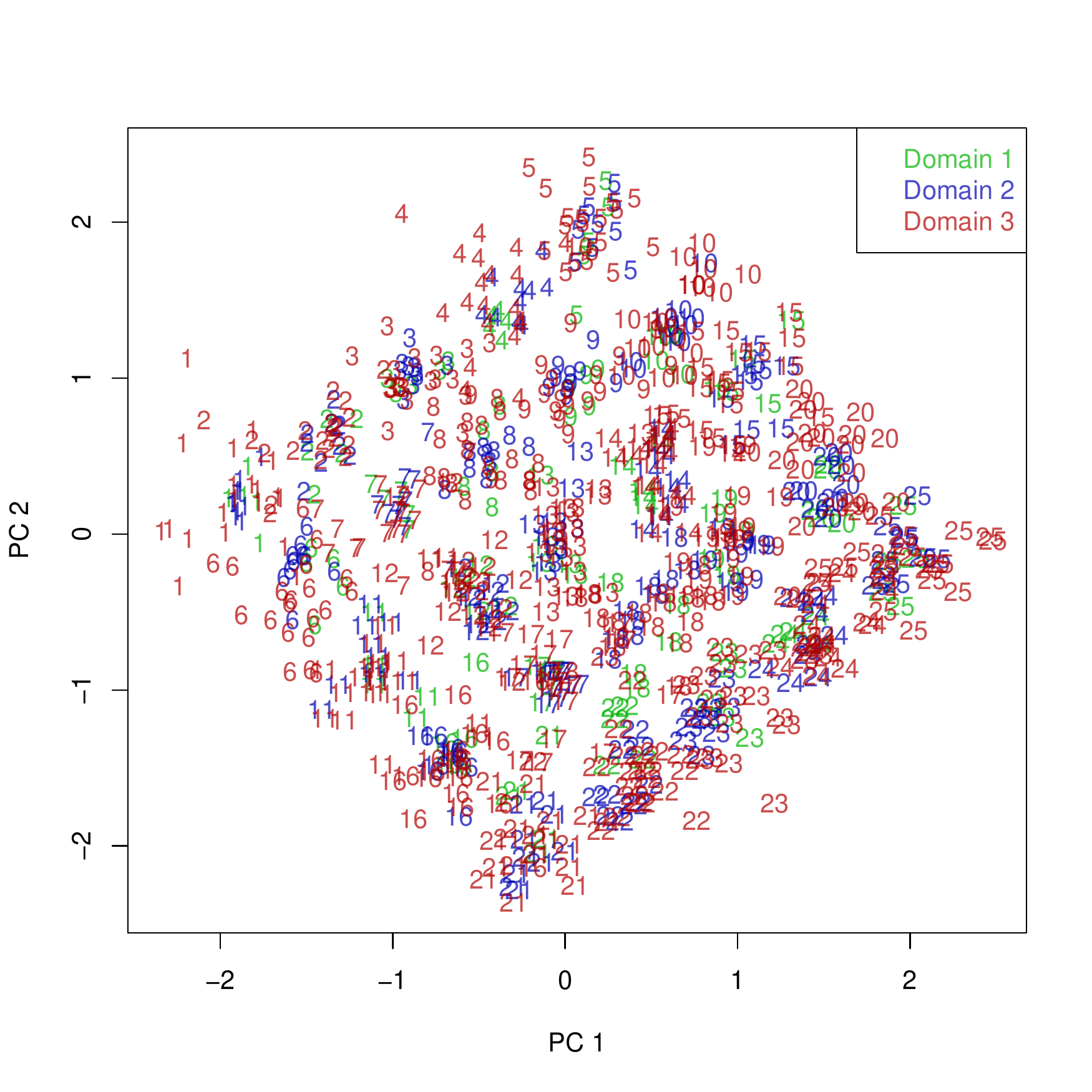}
  \label{fig:g0pc12}
  }
  \subfigure[Cross-validation error of each PC]{
  \includegraphics[width=\figwidth,clip]{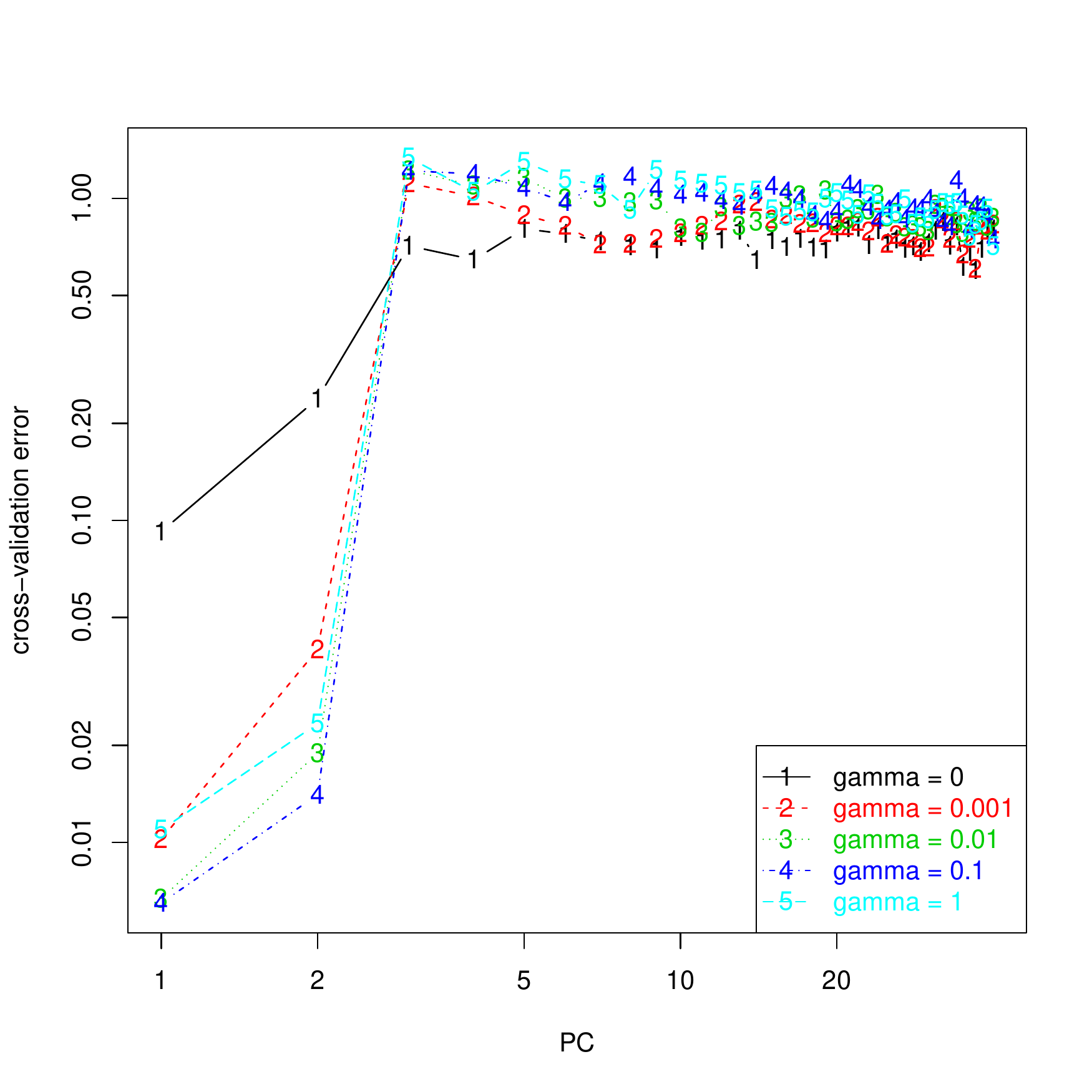}
  \label{fig:crossvalidation}
  }
  \end{center}
 \caption{Cross-validation analysis of
 Section~\ref{sec:crossvalidation}}
  \label{fig:fig3}
\end{figure}

\end{document}